\documentclass[10pt,twocolumn,letterpaper]{article}
\usepackage[pagenumbers]{cvpr}
\usepackage{array}
\usepackage{pifont}
\usepackage{wasysym}
\usepackage{tabularx}
\usepackage{multirow}
\usepackage{placeins}
\usepackage{microtype}
\usepackage{mathtools}
\usepackage{algorithm}
\usepackage{algpseudocode}
\newcommand{\method}{DemoFoundry\xspace}

\newcommand{\E}{\mathbb{E}}
\newcommand{\dataset}{\mathcal{D}}

\newcolumntype{Y}{>{\raggedright\arraybackslash}X}

\definecolor{cvprblue}{rgb}{0.21,0.49,0.74}
\usepackage[breaklinks,colorlinks,allcolors=cvprblue]{hyperref}
\hypersetup{pdftitle={DemoFoundry: Toward Scalable Photorealistic Robot Data with Coding Agents and Generative Rendering},pdfauthor={Ruixiang Wang, Xinyue Yao, Xuanning Zhou, Yueci Deng, Yongyi Su, Kui Jia}}
\title{DemoFoundry: Toward Scalable Photorealistic Robot Data\\with Coding Agents and Generative Rendering}
\author{
{\fontsize{11}{13}\selectfont
Ruixiang Wang$^{1,2,*}$\hspace{9pt}Xinyue Yao$^{1,2,*}$\hspace{9pt}Xuanning Zhou$^{1,*}$\hspace{9pt}Yueci Deng$^{1,2}$\hspace{9pt}Yongyi Su$^{3,\dagger}$\hspace{9pt}Kui Jia$^{1,2}$}\\[7pt]
{\normalsize $^{1}$The Chinese University of Hong Kong, Shenzhen}\\[2pt]
{\normalsize $^{2}$DexForce Technology Co., Ltd.\qquad $^{3}$Foshan University}\\[5pt]
{\small\url{https://robbinw.github.io/DemoFoundry-Web}}
}
\makeatletter
\renewcommand{\@maketitle}{%
  \newpage\null\vspace*{-14pt}%
  \begin{center}
    {\fontsize{16}{19}\selectfont\bfseries\@title\par}\vspace{14pt}
    {\large\lineskip .5em\begin{tabular}[t]{c}\@author\end{tabular}\par}
  \end{center}\vspace{4pt}%
}
\apptocmd{\@maketitle}{\begin{minipage}{\textwidth}
\captionsetup{hypcap=false}
\centering
\includegraphics[width=0.98\textwidth,trim=0 0 0 18,clip]{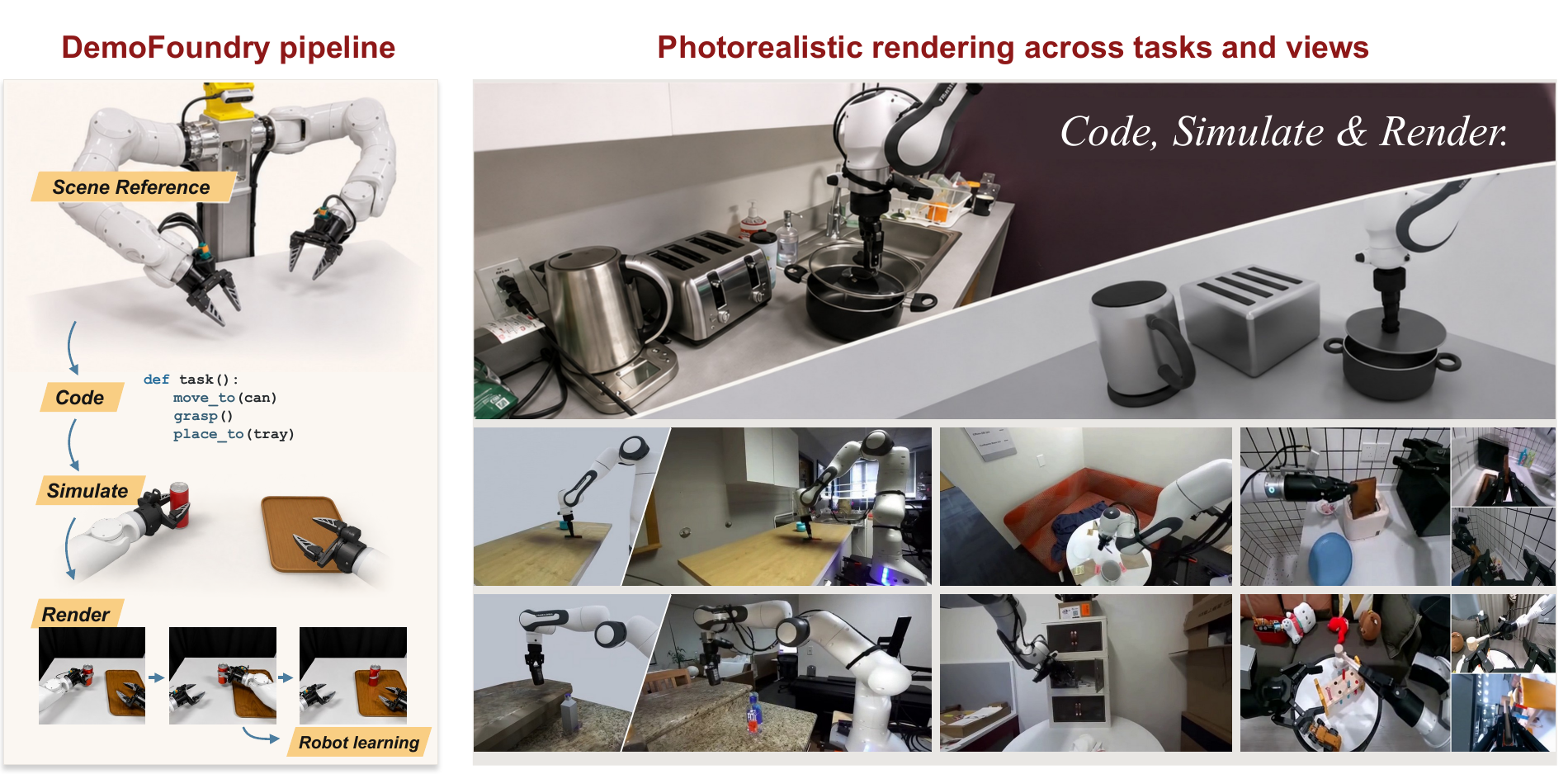}
\captionof{figure}{\textbf{\method{} overview.} The left panel shows how a real-world scene reference guides agent-written programs, physics-based simulation, and video rendering for robot learning. The right panel presents proxy-to-video rendering examples across tasks and camera viewpoints, including external, head, and wrist views.}
\label{fig:overview}
\end{minipage}
\par\vspace{8pt}
}{}{\PackageError{teaser}{Unable to attach first-page teaser}{}}
\makeatother
\newcommand{\FullWidthPage}[1]{\clearpage\twocolumn[{#1}]\null\clearpage}
\begin{document}
\raggedbottom
\setlength{\textfloatsep}{18pt plus 2pt minus 2pt}
\setlength{\dbltextfloatsep}{18pt plus 2pt minus 2pt}
\setlength{\floatsep}{12pt plus 2pt minus 2pt}
\setlength{\dblfloatsep}{12pt plus 2pt minus 2pt}
\captionsetup{skip=9pt}
\maketitle
\begingroup
\renewcommand{\thefootnote}{}
\footnotetext{\textsuperscript{*}Equal contribution.\quad \textsuperscript{\(\dagger\)}Correspondence to: \href{mailto:suyongyi@fosu.edu.cn}{\textcolor{black}{suyongyi@fosu.edu.cn}}.}
\endgroup
\begin{abstract}
Scaling robot learning requires demonstrations that combine realistic visual observations with reliable action supervision. Simulation-based synthesis faces a substantial visual gap relative to real environments, while video-only generation struggles to maintain physically consistent interactions and provide aligned action labels. We present \textbf{DemoFoundry}, an automated framework that couples coding agents with generative rendering to synthesize photorealistic and physically grounded robot demonstrations. Coding agents both construct aligned real--proxy video pairs from existing robot episodes to train a video renderer and write executable interaction programs to generate new demonstrations. Given a scene reference and task instruction, an agent reconstructs task-relevant geometry and writes a control program without requiring target-task human demonstrations. Program execution in a physics simulator produces proxy videos with aligned robot states and actions. The video renderer translates these proxy videos into scene-consistent, photorealistic observations, reducing the need for detailed appearance modeling in each scene. Automated task-execution checks select successful trajectories for rendering and policy learning. Experiments show improved visual quality and action--video consistency. Policies trained on the generated demonstrations outperform raw-simulation training and approach the performance of training on an equal number of real demonstrations.
\end{abstract}

\section{Introduction}
\label{sec:intro}
Generalizable robot policies require large-scale training data covering diverse interactions between robots and their environments. Such data should pair realistic visual observations with reliable action supervision across a broad range of objects, scenes, and tasks. Human teleoperation provides these demonstrations but requires sustained operator effort to execute tasks and reset the environment between trials. Expanding interaction coverage therefore calls for scalable alternatives to manual collection.

Simulation provides a scalable source of robot interactions~\cite{gensim,robogen,gensim2,robotwin}. However, simulated observations still exhibit a substantial visual gap relative to real environments in object appearance, materials, lighting, and background detail. Domain randomization improves robustness to these differences~\cite{domainrand,robotwin}, although matching a particular deployment environment remains challenging. Generative video models can synthesize realistic observations across diverse scenes, but their outputs often exhibit inconsistent contact, object motion, and task execution~\cite{rbench}. Video-only generation also lacks an executable control trace that directly supplies robot actions. Recovering action supervision through inverse dynamics introduces an additional estimation problem~\cite{dreamgen}.

These complementary strengths motivate separating interaction generation from appearance synthesis~\cite{cwm}. Executable programs provide a controllable source of behavior, and video rendering supplies scene appearance. Scaling this approach requires efficient construction of rendering supervision and task programs across scenes. Advances in program-based control and embodied reasoning~\cite{cap,gptpolicy} make coding agents a practical means of coordinating the reconstruction, alignment, and control tools involved in this process.

We introduce \method, an automated data engine combining coding agents and generative rendering, as illustrated in Fig.~\ref{fig:overview}. First, a pair-construction agent reconstructs and aligns existing real robot episodes with proxy videos, forming a paired dataset for training a video renderer. The renderer learns to recover appearance from a scene reference while following robot and object motion. Second, for a novel scene, an agent reconstructs task-relevant geometry and develops a program to execute the instructed task. Executing this program under different object layouts in simulation produces proxy videos with aligned robot states and actions. The renderer then converts these proxy videos into photorealistic observations using the corresponding scene references.

\method{} enables zero-shot data generation for novel robot embodiments and scenes from reference images and robot/camera configurations, without target-scene demonstrations or human operator involvement during generation. We evaluate it on a real-world platform across three manipulation tasks using Diffusion Policy and $\pi_{0.5}$~\cite{diffusionpolicy,pi05}. Our renderers improve visual quality over baselines while maintaining strong action--video consistency. Policies trained on our generated data outperform raw-simulation training and approach the performance of training on an equal number of real demonstrations. Policy performance improves with more synthetic training data, and estimated generation costs fall below human teleoperation at scale.

Table~\ref{tab:related_comparison} places these components within existing robot data-generation pipelines. We make three contributions.
\begin{itemize}
\item An automated data engine that couples coding-agent programs with appearance rendering to synthesize photorealistic, action-labeled robot demonstrations across diverse scenes and tasks without manual intervention during production.
\item An agent-constructed dataset of 1,884 aligned real--proxy video pairs from four public robot datasets, used to adapt two video-generation models for rendering newly programmed interactions. We plan to release the paired dataset upon publication.
\item Real-robot experiments demonstrating improved policy learning over raw simulation and performance comparable to real-demonstration training, complemented by visual-quality comparisons, paired-training ablation, and data-scaling experiments.
\end{itemize}

\section{Related Work}
\label{sec:related}
\begin{figure*}[!t]
\centering
\includegraphics[width=0.98\textwidth]{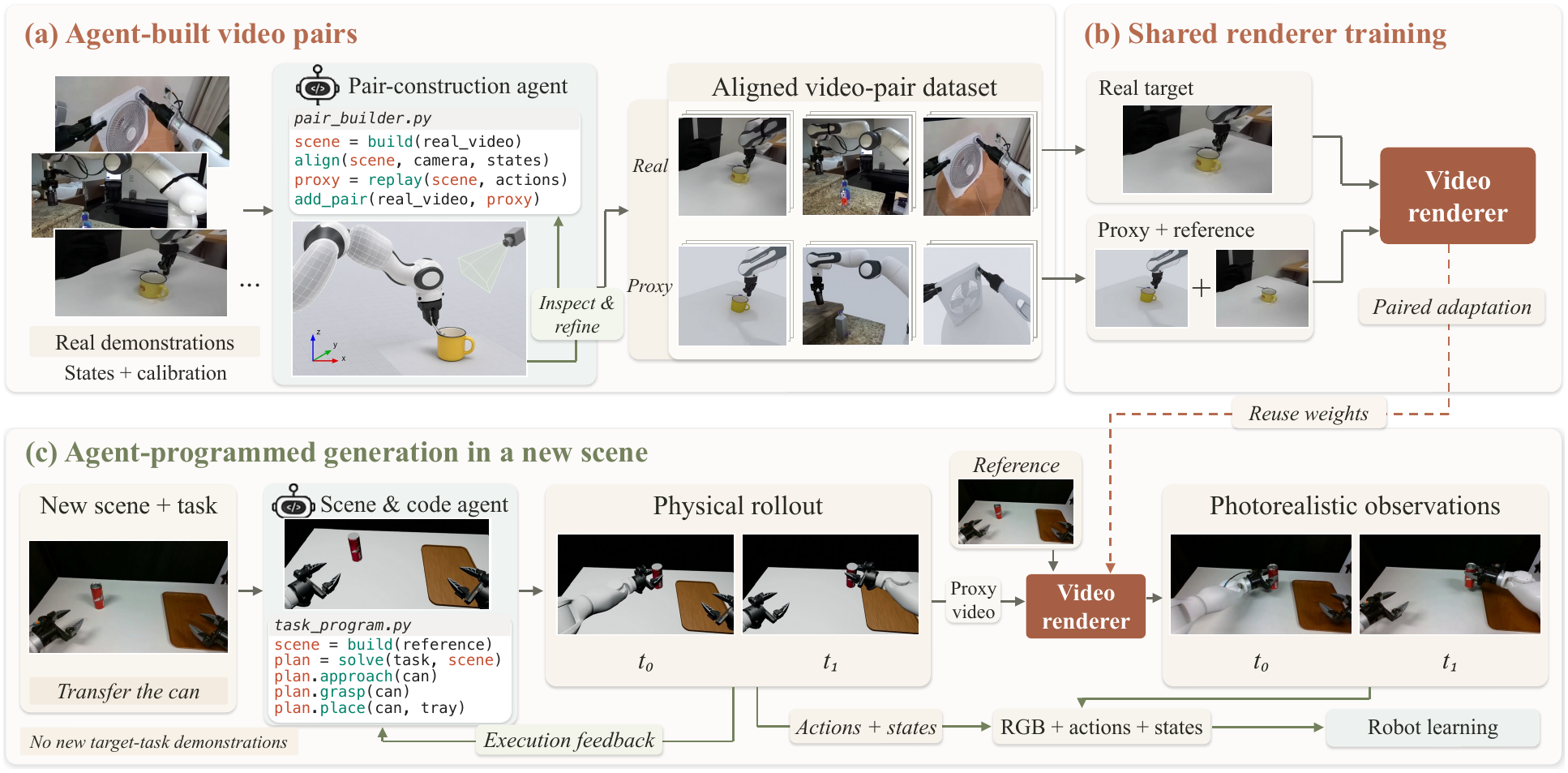}
\caption{\textbf{DemoFoundry pipeline.} (a) A pair-construction agent writes and refines reconstruction, alignment, and simulation programs to convert diverse existing demonstrations into real--proxy video pairs. (b) The paired dataset supervises a video renderer conditioned on a proxy video and an initial RGB reference frame. (c) In a new scene, an agent builds a task program and executes it under randomized simulator layouts. For successful executions, initial proxy frames guide matching RGB-reference generation before video rendering. Simulation retains actions and robot states for robot learning without new target-task demonstrations.}
\label{fig:method}
\end{figure*}

\begin{table}[t]
\centering\fontsize{8}{9.3}\selectfont
\setlength{\tabcolsep}{1.2pt}
\renewcommand{\arraystretch}{1.18}
\newcommand{\capyes}{\textcolor[HTML]{337E63}{\ding{51}}}
\newcommand{\capno}{\textcolor[HTML]{B35A53}{\ding{55}}}
\caption{\textbf{Robot data-generation mechanisms.} \capyes\ included, \capno\ absent in the reported workflow. Paired data combines real RGB videos with simulated episode replays, labeled Replay, or extracted geometric controls, labeled Derived, with --- indicating absent pairing. Zero-shot denotes setting up the generator without target-scene demonstrations, given scene images or scans and robot/camera configurations.}
\label{tab:related_comparison}
\begin{tabularx}{\columnwidth}{@{}p{0.29\columnwidth}>{\centering\arraybackslash}p{0.13\columnwidth}*{5}{>{\centering\arraybackslash}X}@{}}
\toprule
\multirow{2}{*}{Method} & Paired & Zero- & Scene & Task & Sim. & Video \\
& data & shot & agent & code & acts & model \\
\midrule
RoboTwin 2.0~\cite{robotwin} & --- & \capyes & \capno & \capyes & \capyes & \capno \\
Re$^3$Sim~\cite{re3sim} & --- & \capyes & \capno & \capno & \capyes & \capno \\
Agentic Real2Sim~\cite{agenticreal2sim} & Replay & \capno & \capyes & \capno & \capyes & \capno \\
DreamGen~\cite{dreamgen} & --- & \capyes & \capno & \capno & \capno & \capyes \\
RoboRender~\cite{roborender} & Derived & \capyes & \capno & \capno & \capyes & \capyes \\
ComSim~\cite{comsim} & Replay & \capno & \capno & \capyes & \capyes & \capyes \\
DataFarm~\cite{datafarm} & --- & \capno & \capyes & \capyes & \capyes & \capyes \\
\midrule
\textbf{DemoFoundry} & Replay & \capyes & \capyes & \capyes & \capyes & \capyes \\
\bottomrule
\end{tabularx}
\end{table}

\subsection{Programmatic Simulation and Data Engines}
Programmatic simulation scales robot demonstrations through automatic task specification and behavior generation. Language-model-based systems synthesize executable policies, task programs, and learning objectives~\cite{cap,gensim,robogen,gensim2}. RoboTwin~2.0 combines feedback-driven expert-code generation with domain randomization for bimanual manipulation~\cite{robotwin}, while Genie Sim~3.0 provides high-fidelity simulation infrastructure and diverse assets~\cite{geniesim}. \method extends this paradigm by coupling agent-generated behavior with a video renderer trained on aligned real--proxy observations.

\subsection{Robot Data Synthesis from Real Observations}
Real observations support data generation through trajectory reuse and scene reconstruction. Demonstration-amplification methods transform recorded motion into new object configurations~\cite{mimicgen,dexmimicgen,demogen}, while replay and reconstructed-asset rendering expand visual and embodiment coverage~\cite{rebot,real2render2real}. Digital-twin methods recover scene appearance and geometry for policy learning~\cite{rialto,splatsim,re3sim}, and IGen generates visuomotor trajectories from open-world images~\cite{igen}. Agentic Real2Sim automates episode reconstruction and physical replay, producing real--simulation correspondences~\cite{agenticreal2sim}. \method uses such correspondences across multiple robot datasets to train a reusable video renderer, then generates new-task trajectories from scene references and task programs.

\subsection{Generative Rendering for Robot Learning}
Generative models provide photorealistic observations for robot learning. DreamGen and EVA synthesize behavior videos and recover action supervision through inverse dynamics~\cite{dreamgen,eva}. Trajectory-conditioned approaches instead guide appearance synthesis with robot motion, edited geometry, or simulator controls~\cite{anchordream,robodream,real2edit2real,craft}. RoboTransfer uses multi-view geometric conditions~\cite{robotransfer}, and RoboRender trains on real-derived depth and robot-mask controls paired with RGB videos~\cite{roborender}. Code World Model and SolarRenderer provide related program-driven and structural-video rendering mechanisms~\cite{cwm,solarrenderer}. Hybrid robot data engines combine executable programs with video rendering. ComSim~\cite{comsim} constructs real--simulation pairs by replaying recorded actions, while PyRUA-DataFarm~\cite{datafarm} builds its target scene from one calibration episode. \method connects agent-built, multi-source real--proxy pairs to renderer adaptation and new-scene demonstration generation.

\section{DemoFoundry}
\label{sec:method}
\subsection{Overview and Formulation}
As illustrated in Fig.~\ref{fig:method}, DemoFoundry first constructs aligned real--proxy video pairs to train a video renderer, as detailed in Sec.~\ref{sec:pairs}. To generate demonstrations in a new scene, a coding agent takes a reference image $I^{\mathrm{ref}}$, a robot specification $R$ defining the robot and gripper models, camera calibration $C$ specifying intrinsics and extrinsics, and a task instruction $\ell$ to build a simulated scene $E$ and task program $p$.

Executing $p$ in $E$ yields
\begin{equation}
(P,Q,A)=\operatorname{Execute}(E,p;C),
\label{eq:execution}
\end{equation}
where $P$ is the proxy video, $Q$ the aligned robot state sequence, and $A$ the executed actions. The trained video renderer synthesizes photorealistic observations conditioned on the proxy and reference image,
\begin{equation}
\widehat V\sim G_{\theta}(\cdot\mid P,I^{\mathrm{ref}}).
\label{eq:renderer}
\end{equation}
The rendered video retains the simulator's state and action labels, forming demonstrations $(\widehat V,Q,A,\ell)$ for the policy-training dataset $\dataset_{\mathrm{gen}}$.

\subsection{Video Renderer Training}
\label{sec:pairs}
\begin{figure}[t]
\centering
\includegraphics[width=\columnwidth]{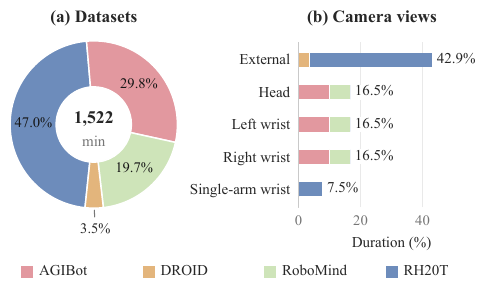}
\caption{\textbf{Composition of the paired-video dataset.} Video duration by (a) source dataset and (b) camera view. Colors identify source datasets, and durations are summed across camera views.}
\label{fig:dataset_composition}
\end{figure}

\paragraph{Agent-driven reconstruction.}
We construct real--proxy video pairs from DROID~\cite{droid}, AgiBot World~\cite{agibot}, RoboMIND~\cite{robomind}, and RH20T~\cite{rh20t} to supervise video renderer training. For each source episode, a GPT-6 Astra agent~\cite{gpt6astra} reconstructs the task-relevant scene from RGB observations, available depth, camera calibration, and robot/gripper records. It obtains object masks with SAM~3.1~\cite{sam3} and fits the tabletop as the support surface. Simple objects are represented by parameterized shapes and converted into meshes through agent-written Blender Python scripts~\cite{scenecraft,ll3m}. Complex assets are reconstructed from RGB images and masks using SAM 3D Objects~\cite{sam3dobjects}. Available depth initializes scale and placement, which the agent refines using multi-view silhouettes, object orientation, and support constraints. The resulting visual and collision geometry preserves task-relevant openings and contact surfaces for physical replay.

\paragraph{Physical execution and refinement.}
The reconstructed assets and corresponding robot/gripper models are assembled in a physics engine, such as MuJoCo~\cite{mujoco} or DexSim~\cite{embodichain}. The agent replays the recorded actions through the robot controller, with contact dynamics determining object motion. Rendering the simulated trajectories at the recorded camera poses and timestamps yields candidate proxy videos. To align each proxy with its real-video counterpart, we compare object masks at sampled keyframes and require silhouette IoU of at least 0.80 for every visible object in every view. Failed comparisons guide adjustments to object geometry, scale, or initial placement, followed by another replay with calibration fixed. We allow up to three refinement rounds after the initial replay and discard pairs that still fail. Accepted pairs retain camera calibration, scene assets, timestamps, and robot state/action traces. Appendix A describes the agent implementation and pair-construction evaluation.

This procedure produces 1,884 aligned real--proxy video pairs totaling 1,522.03 minutes summed across camera views, with the dataset breakdown shown in Fig.~\ref{fig:dataset_composition}. We plan to release the dataset upon publication to support training video models for sim-to-real transfer and photorealistic robot-video rendering. The paired simulations also provide object poses and scene states, offering potential supervision for 3D VLA models~\cite{3dvla} and 3D world models such as PointWorld~\cite{pointworld} and UNITAS~\cite{unitas}.

\paragraph{Conditional renderer training.}
The accepted pairs form the training dataset
\begin{equation}
\dataset_{\mathrm{pair}}=\{(V_i,P_i,I_i^{\mathrm{ref}})\}_{i=1}^{M},
\end{equation}
where $V_i$ is the real target video, $P_i$ its aligned proxy, and $I_i^{\mathrm{ref}}=V_{i,1}$ its initial RGB frame. We adapt a pretrained video model using $P_i$ and $I_i^{\mathrm{ref}}$ as conditioning inputs and $V_i$ as supervision. For conditional flow matching, let $z=\mathcal E(V)$ be the encoded target, $\epsilon\sim\mathcal N(0,I)$, and $z_u=(1-u)\epsilon+uz$, with $u\sim\rho$ sampled according to the backbone's training schedule. We optimize
\begin{equation}
\mathcal L_{\mathrm{FM}}=
\E\left[\left\|v_\theta^{\mathrm{train}}(z_u,u;P,I^{\mathrm{ref}})-(z-\epsilon)\right\|_{\mathrm{valid}}^2\right],
\label{eq:training}
\end{equation}
where $v_\theta^{\mathrm{train}}$ predicts the target velocity and the loss is averaged over valid target-latent elements. The trained renderer $G_\theta$ is then reused to render newly executed interactions in Sec.~\ref{sec:engine}.

\subsection{Demonstration Generation in New Scenes}
\label{sec:engine}
\begin{algorithm}[t]
\caption{Demonstration generation in a new scene}
\label{alg:generation}
\small
\algrenewcommand\algorithmicindent{1em}
\renewcommand{\algorithmicrequire}{\textbf{Input:}}
\renewcommand{\algorithmicensure}{\textbf{Output:}}
\begin{algorithmic}[1]
\Require Scene reference image $I^{\mathrm{ref}}$, robot specification $R$, camera calibration $C$, task instruction $\ell$, target demonstration count $N$, tool library $\mathcal T$, image editor $\mathcal M_{\mathrm{img}}$, video renderer $G_\theta$
\Ensure Demonstrations $\dataset_{\mathrm{gen}}=\{(\widehat V_i,Q_i,A_i,\ell)\}$
\Statex \textbf{Agentic task programming}
\State Reconstruct task-relevant assets from $I^{\mathrm{ref}}$ as in Sec.~\ref{sec:pairs}; assemble $E$ with robot $R$ and cameras $C$
\State Agent writes pose-aware program $p$ for $\ell$ using $\mathcal T$; trial execution guides code corrections
\State $\dataset_{\mathrm{gen}}\gets\varnothing$; return if $p$ fails the task
\Statex \textbf{Demonstration rendering}
\While{$|\dataset_{\mathrm{gen}}|<N$}
  \State $i\gets|\dataset_{\mathrm{gen}}|+1$
  \State Randomize object positions in $E$ to construct $E_i$
  \State $(P_i,Q_i,A_i)\gets\operatorname{Execute}(E_i,p;C)$
  \State \textbf{if} task execution fails \textbf{then continue}
  \State $I_i^{\mathrm{ref}}\gets\mathcal M_{\mathrm{img}}(P_{i,1},I^{\mathrm{ref}})$ \hfill (layout, appearance)
  \State Render $P_i$ in windows with $G_\theta$, using $I_i^{\mathrm{ref}}$ initially and the previous output's last frame thereafter
  \State Concatenate rendered segments into $\widehat V_i$
  \State $\dataset_{\mathrm{gen}}\gets\dataset_{\mathrm{gen}}\cup\{(\widehat V_i,Q_i,A_i,\ell)\}$
\EndWhile
\State \Return $\dataset_{\mathrm{gen}}$
\end{algorithmic}
\end{algorithm}

\paragraph{Agentic task programming.}
Given a new scene reference $I^{\mathrm{ref}}$, robot specification $R$, camera calibration $C$, and task instruction $\ell$, a GPT-6 Astra agent reconstructs task-relevant assets using the procedure in Sec.~\ref{sec:pairs}. It imports the robot/gripper models and assembles the simulated scene $E$. Following code-based robot control~\cite{cap} and simulation data engines~\cite{gensim,robotwin}, the agent writes an executable task program $p$ using a tool library $\mathcal T$ for scene queries, grasp proposals, inverse kinematics, motion planning, and robot control. Trial execution checks task completion and guides code corrections. The program queries current object poses, allowing it to be reused across different layouts.

\paragraph{Demonstration rendering.}
To generate diverse demonstrations, we randomize object positions in the reconstructed simulator scene $E$ to obtain $E_i$ and execute $p$ under camera calibration $C$, recording the proxy video $P_i$, robot states $Q_i$, and actions $A_i$. For each successful execution, the image-generation model $\mathcal M_{\mathrm{img}}$ uses the first proxy frame $P_{i,1}$ for layout and the original $I^{\mathrm{ref}}$ for appearance to generate a matching photorealistic reference $I_i^{\mathrm{ref}}$. The trained renderer $G_\theta$ then synthesizes photorealistic video $\widehat V_i$ conditioned on $P_i$ and $I_i^{\mathrm{ref}}$. Each camera view uses its corresponding proxy and reference, with shared simulation timestamps and state/action labels.

For long trajectories, rendering proceeds autoregressively in windows of the model's training clip length. The first window uses $I_i^{\mathrm{ref}}$ and the initial proxy segment. Subsequent windows use the preceding output's final frame and the next proxy segment, with a one-frame overlap. We concatenate the segments, keeping the shared boundary frame once, and pair the resulting video $\widehat V_i$ with the executed states and actions to form $(\widehat V_i,Q_i,A_i,\ell)$. Algorithm~\ref{alg:generation} repeats this process until the target number $N$ of demonstrations has been collected in $\dataset_{\mathrm{gen}}$.

\paragraph{Robot policy learning.}
We use $\dataset_{\mathrm{gen}}$ to train Diffusion Policy~\cite{diffusionpolicy} and $\pi_{0.5}$~\cite{pi05}. Both receive rendered observations and robot states as inputs, with aligned simulator actions as supervision and task instruction $\ell$ additionally supplied to $\pi_{0.5}$. At deployment, the policies receive real camera observations and robot states. Section~\ref{sec:implementation} and Appendix B provide training and deployment settings.

\section{Experiments}
\label{sec:experiments}
\begin{figure*}[!t]
\centering
\includegraphics[width=0.95\textwidth,trim=0 2 0 6,clip]{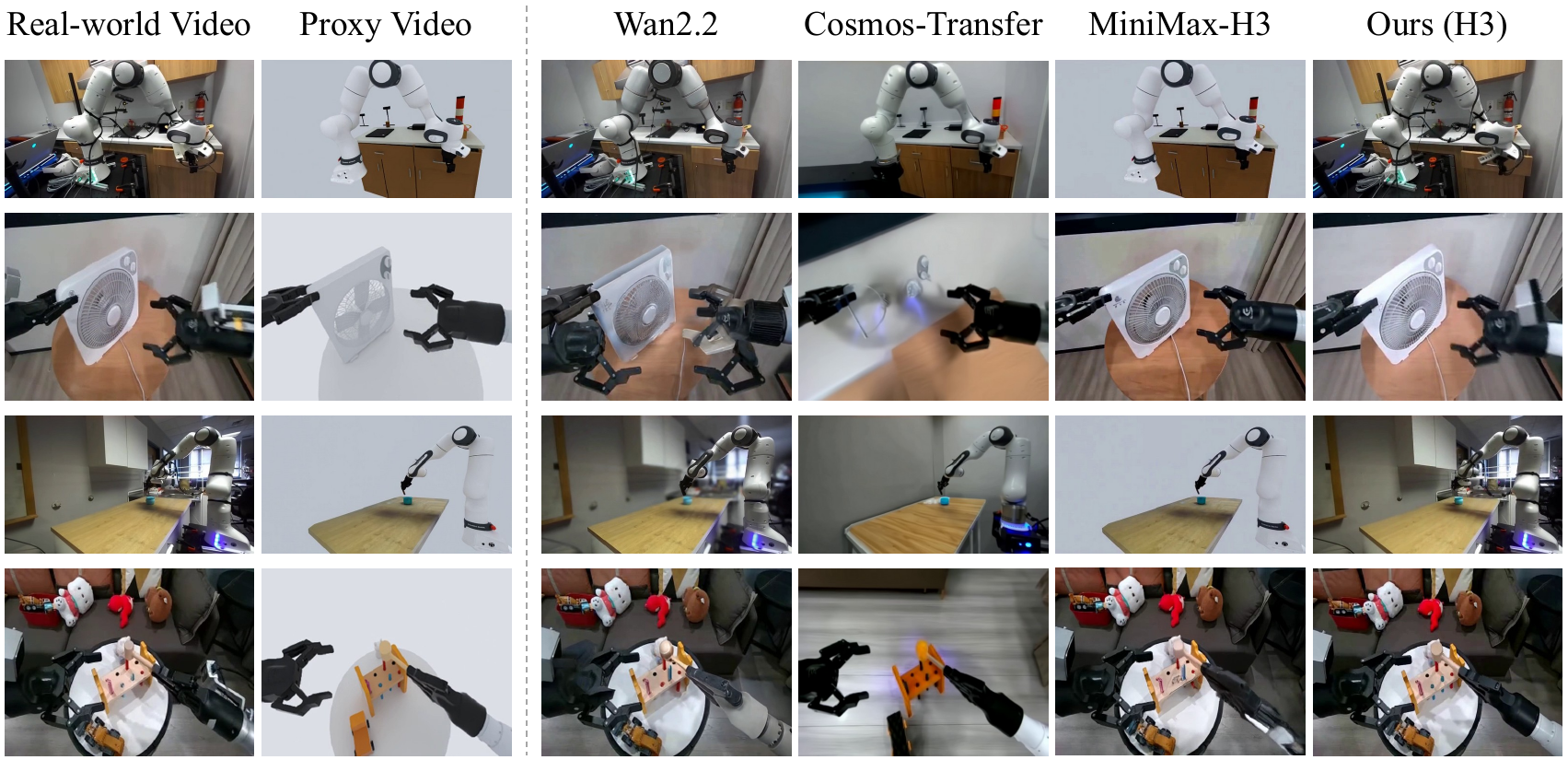}
\caption{\textbf{Qualitative comparison of robot video rendering.} Each row shows corresponding frames from a real video, its proxy, and the compared rendering methods.}
\label{fig:qualitative}
\end{figure*}

\begin{table*}[t]
\centering\small
\setlength{\tabcolsep}{5pt}
\caption{\textbf{Visual quality on held-out real--proxy pairs.} Methods use the same trajectories and evaluation timestamps through their native interfaces. The LTX and H3 baselines use released weights. Our variants are fine-tuned on the same paired dataset. Best results in each column are bold.}
\label{tab:video}
\begin{tabularx}{\textwidth}{@{}Yccccc@{}}
\toprule
Method & PSNR (dB)$\uparrow$ & SSIM~\cite{ssim}$\uparrow$ & LPIPS~\cite{lpips}$\downarrow$ & FID~\cite{fid}$\downarrow$ & FVD~\cite{fvd}$\downarrow$ \\
\midrule
Raw simulation & 8.468 & 0.5005 & 0.6272 & 237.48 & 1618.89 \\
Wan2.2-Fun-5B-Control~\cite{wanfun} & 14.537 & 0.6548 & 0.3412 & 155.26 & 897.33 \\
Cosmos-Transfer2.5~\cite{cosmos25} & 12.115 & 0.4929 & 0.5769 & 229.89 & 1290.42 \\
RoboTransfer~\cite{robotransfer} & 15.008 & 0.6287 & 0.3614 & 161.73 & 1018.46 \\
RoboRender~\cite{roborender} & 16.925 & 0.6612 & 0.2949 & 116.81 & 815.21 \\
LTX-2.5 Layout-to-Render~\cite{ltx25layout} & 16.989 & 0.6552 & 0.3014 & 127.43 & 886.74 \\
MiniMax-H3 Ref2VA~\cite{minimaxh3} & 15.462 & 0.5281 & 0.3294 & 121.01 & 808.74 \\
\midrule
\textbf{Ours (LTX)}~\cite{ltx25layout} & 22.092 & 0.8072 & 0.1682 & 74.42 & 454.62 \\
\textbf{Ours (H3)}~\cite{minimaxh3} & \textbf{23.317} & \textbf{0.8569} & \textbf{0.1242} & \textbf{52.88} & \textbf{258.81} \\
\bottomrule
\end{tabularx}
\end{table*}
\begin{table}[t]
\centering\small
\setlength{\tabcolsep}{3pt}
\renewcommand{\arraystretch}{1.2}
\caption{\textbf{Action--video consistency.} MiniMax-H3 uses released Ref2VA weights. PCK and EventAlign are percentages. Best results are bold.}
\label{tab:action_video}
\begin{tabular*}{\columnwidth}{@{\extracolsep{\fill}}lcccc@{}}
\toprule
Method & PCK@5\%$\uparrow$ & \shortstack{Flow\\nEPE$\downarrow$} & \shortstack{TRAJAN\\distance$\downarrow$} & \shortstack{EventAlign\\$\pm0.5$\,s$\uparrow$} \\
\midrule
RoboRender~\cite{roborender} & 86.57 & 0.456 & 26.060 & 54.55 \\
MiniMax-H3~\cite{minimaxh3} & 93.55 & 0.461 & 24.251 & 80.95 \\
\textbf{Ours (H3)}~\cite{minimaxh3} & \textbf{97.24} & \textbf{0.433} & \textbf{22.294} & \textbf{95.23} \\
\bottomrule
\end{tabular*}
\end{table}
\begin{figure*}[!t]
\centering
\includegraphics[width=\textwidth]{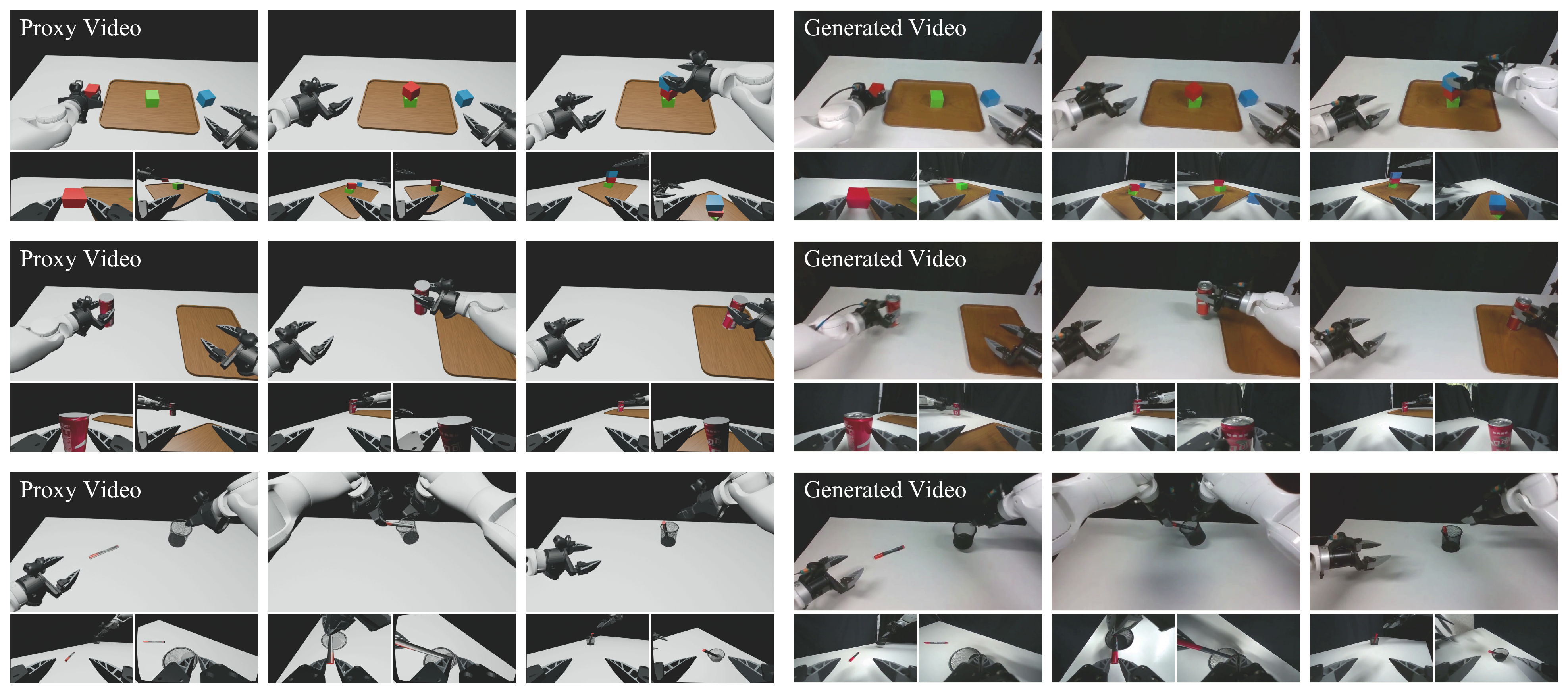}
\caption{\textbf{Photorealistic demonstration synthesis on Marvin M6S.} Rows show block stacking, can-to-tray placement, and pen insertion. Each row compares three corresponding time points from proxy videos on the left and generated videos on the right, with head views above paired wrist views.}
\label{fig:unseen_platform}
\end{figure*}

\begin{table*}[t]
\centering\small
\setlength{\tabcolsep}{3pt}
\renewcommand{\arraystretch}{1.12}
\caption{\textbf{Real-robot learning results.} Task entries are successes out of 30 trials for Diffusion Policy (DP)~\cite{diffusionpolicy} and $\pi_{0.5}$~\cite{pi05}. Average reports the three-task mean success rate (\%). Real/Syn. are training-demonstration counts per task. Best results are bold.}
\label{tab:policy}
\begin{tabularx}{\textwidth}{@{}p{0.23\textwidth}rr*{2}{>{\centering\arraybackslash}X}@{\hspace{9pt}}*{2}{>{\centering\arraybackslash}X}@{\hspace{9pt}}*{2}{>{\centering\arraybackslash}X}@{\hspace{9pt}}*{2}{>{\centering\arraybackslash}X}@{}}
\toprule
& & & \multicolumn{2}{c}{Can to tray} & \multicolumn{2}{c}{Block stacking} & \multicolumn{2}{c}{Pen to holder} & \multicolumn{2}{c}{Average (\%)$\uparrow$} \\
\cmidrule(lr){4-5}\cmidrule(lr){6-7}\cmidrule(lr){8-9}\cmidrule(l){10-11}
Training data & Real & Syn. & DP & $\pi_{0.5}$ & DP & $\pi_{0.5}$ & DP & $\pi_{0.5}$ & DP & $\pi_{0.5}$ \\
\midrule
Simulation only & 0 & 50 & 7/30 & 11/30 & 4/30 & 8/30 & 0/30 & 6/30 & 12.2 & 27.8 \\
Real only & 50 & 0 & 18/30 & 23/30 & 13/30 & 16/30 & 13/30 & 17/30 & 48.9 & 62.2 \\
Real + Simulation & 50 & 100 & 17/30 & 22/30 & 15/30 & 19/30 & 11/30 & 16/30 & 47.8 & 63.3 \\
\midrule
\textbf{DemoFoundry only} & 0 & 50 & 17/30 & 22/30 & 13/30 & 17/30 & 12/30 & 15/30 & 46.7 & 60.0 \\
\textbf{Real + DemoFoundry} & 50 & 100 & \textbf{24/30} & \textbf{27/30} & \textbf{21/30} & \textbf{23/30} & \textbf{20/30} & \textbf{24/30} & \textbf{72.2} & \textbf{82.2} \\
\bottomrule
\end{tabularx}
\end{table*}

\subsection{Implementation Details}
\label{sec:implementation}
\paragraph{Video renderer training.}
We train two video renderers $G_\theta$ based on LTX-2.5 Layout-to-Render~\cite{ltx25layout} and MiniMax-H3 Ref2VA~\cite{minimaxh3}, denoted \textbf{Ours (LTX)} and \textbf{Ours (H3)}, respectively. Both are fine-tuned on our aligned real--proxy pairs with source-balanced sampling while keeping the backbones and visual encoders frozen. LTX continues training the released rank-128 LoRA on 121-frame clips. H3 uses rank-32 attention LoRA on 158-frame clips at 24\,fps, with a frozen DeCFG correction~\cite{decfg} enabled during training and removed at inference. Training resolutions are $512\times384$ for AgiBot/RoboMIND and $640\times352$ for DROID/RH20T, expressed as width$\times$height. Appendix B lists the training configurations.

\paragraph{Inference settings.}
The autoregressive rollout in Sec.~\ref{sec:engine} uses 121-frame windows for LTX and 158-frame windows for H3. LTX uses 30 denoising steps with guidance scale 3, while H3 uses 50 steps with its native sampler. We use Qwen-Image-2.1~\cite{qwenimage21} for reference-image editing, with GPT Image~2.5~\cite{gptimage25} as an alternative. The H3 rendering prompt, detailed in Appendix B, describes scene appearance and instructs the model to follow proxy motion. Appearance edits modify this prompt, whereas task instruction $\ell$ guides the agent and policy.

\paragraph{Policy training.}
We use our synthetic demonstrations to train two downstream robot policies, Diffusion Policy (DP)~\cite{diffusionpolicy} and $\pi_{0.5}$~\cite{pi05}. We fine-tune $\pi_{0.5}$ from the pretrained $\pi_{0.5}$-base weights for 20,000 steps with a batch size of 32, and train DP for 50,000 steps with a batch size of 64. On the real robot, $\pi_{0.5}$ receives three camera views and joint/gripper states; both policies run at 25\,Hz. Appendix B gives the action representation and optimization settings.

\subsection{Visual Fidelity}
\label{sec:visualexp}

\paragraph{Baselines.}
\textbf{Raw simulation} directly uses the proxy video. General conditional baselines include \textbf{Wan2.2-Fun-5B-Control}~\cite{wanfun} with structural-video controls, \textbf{LTX-2.5 Layout-to-Render}~\cite{ltx25layout} with layout/reference conditioning, and \textbf{MiniMax-H3 Ref2VA}~\cite{minimaxh3} with image/video references. \textbf{Cosmos-Transfer2.5}~\cite{cosmos25} provides spatially controlled sim-to-real appearance transfer. \textbf{RoboTransfer}~\cite{robotransfer} uses depth, normals, and appearance references, and \textbf{RoboRender}~\cite{roborender} uses depth, robot RGB masks, text, and previous-frame conditioning. Required geometry controls come from the same simulated trajectories. Released and adapted LTX/H3 variants use matched inputs and sampling settings.

\paragraph{Quantitative results.}
We evaluate rendering quality on held-out real--proxy pairs using peak signal-to-noise ratio (PSNR) for pixel agreement, SSIM~\cite{ssim} for structural similarity, LPIPS~\cite{lpips} for perceptual discrepancy, and FID~\cite{fid} and FVD~\cite{fvd} for image- and video-distribution similarity. As reported in Table~\ref{tab:video}, both adapted renderers outperform all baselines across all five metrics, with \textbf{Ours (H3)} performing best. Relative to the released LTX and H3 backbones, paired fine-tuning increases PSNR by 5.10 and 7.86\,dB and reduces FVD by 48.7\% and 68.0\%, respectively. Figure~\ref{fig:qualitative} illustrates recovered scene appearance across robot manipulation tasks.

\paragraph{Action--video consistency.}
To assess consistency with simulator action labels, we compare rendered and proxy videos at matched timestamps. We measure gripper/object localization with semantic PCK@5\%, optical-flow agreement with normalized RAFT~\cite{raft} endpoint error (nEPE), trajectory similarity with TRAJAN~\cite{trajan} embedding distance from BootsTAPIR~\cite{bootstap} tracks, and interaction timing with EventAlign@$0.5$\,s. Event alignment requires matching object identity and a full annotated time-difference interval within $\pm0.5$\,s. As shown in Table~\ref{tab:action_video}, our H3 renderer performs best across all four metrics, improving EventAlign from 80.95\% to 95.23\% over the released H3 model.

\subsection{Real-Robot Policy Learning}
\label{sec:policyexp}

We evaluate robot policy learning on a TianJi Marvin M6S (CCS-696) dual-arm platform~\cite{marvin}, a new embodiment for our demonstration-generation pipeline shown in Fig.~\ref{fig:unseen_platform}. Unless otherwise specified, subsequent experiments use our fine-tuned H3 renderer for data generation. We evaluate three real-world manipulation tasks: placing a can into a tray, stacking blocks, and inserting a pen into a pen holder. We train Diffusion Policy~\cite{diffusionpolicy} as a visuomotor baseline and $\pi_{0.5}$~\cite{pi05} as the VLA policy separately for each task. Success requires the specified placement or stack to remain stable after release.

\begin{figure*}[!t]
\begin{minipage}[t]{\columnwidth}
\vspace{0pt}
\centering
\includegraphics[width=\columnwidth]{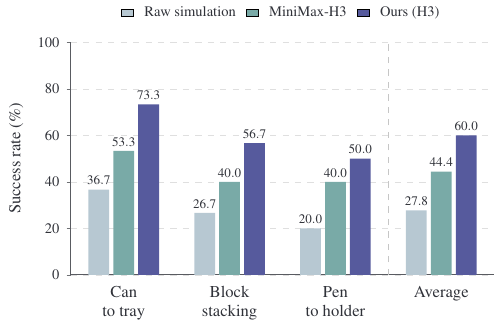}
\caption{\textbf{Paired renderer training improves policy transfer.} $\pi_{0.5}$ success rates over 30 real-robot trials per task. Average gives equal weight to the three tasks.}
\label{fig:paired_renderer}
\end{minipage}\hfill
\begin{minipage}[t]{\columnwidth}
\vspace{0pt}
\centering
\includegraphics[width=\columnwidth]{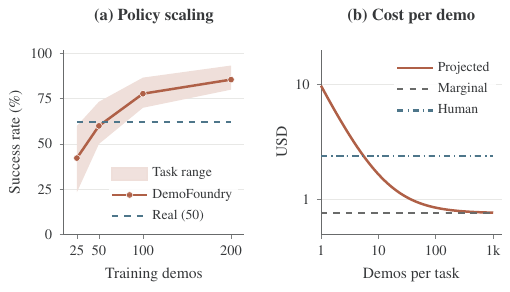}
\caption{\textbf{Data scaling and projected generation cost.} (a) Mean $\pi_{0.5}$ success across three tasks, with their minimum-to-maximum range shaded. The dashed line uses 50 real demonstrations per task. (b) Estimated generation cost per 30-s, three-view demonstration, with a one-time planning cost of approximately \$9 per task and an assumed image/video GPU cost of \$0.765 per demonstration. The human teleoperation reference is \$2.42 per demonstration.}
\label{fig:scaling}
\end{minipage}
\end{figure*}

Table~\ref{tab:policy} compares five data conditions with 30 real-robot trials per task and policy. Simulation-only and DemoFoundry-only share 50 trajectories per task, differing only in visual observations; real-only uses 50 human demonstrations. Mixed conditions add 100 raw or H3-rendered trajectories to the same real data. We report success counts and mean success across tasks. DemoFoundry-only training improves mean success over raw simulation by 34.4 percentage points for DP and 32.2 for $\pi_{0.5}$. Adding 100 raw-simulation trajectories has little effect on mean success relative to real-only training, changing it by $-1.1$ percentage points for DP and $+1.1$ for $\pi_{0.5}$. Adding generated demonstrations to real data further improves over real-only training by 23.3 and 20.0 percentage points, respectively.

\subsection{Ablation and Data Scaling}
\label{sec:ablations}
\paragraph{Ablation on video renderer training.}
Figure~\ref{fig:paired_renderer} compares raw simulation, released H3, and adapted H3 with the same 50 trajectories per task and fixed policy training settings. H3 variants share conditioning inputs and sampling settings. Paired fine-tuning raises mean success by 15.6 percentage points, from 44.4\% to 60.0\%. All three tasks benefit from paired renderer training.

\paragraph{Policy data scaling.}
Figure~\ref{fig:scaling}(a) evaluates $\pi_{0.5}$ with 25, 50, 100, and 200 generated trajectories per task, holding the renderer, policy initialization, and optimization budget fixed. Each scale uses 30 trials per task; the curve reports their mean success, with the task minimum--maximum range shaded. Mean success increases from 42.2\% at 25 demonstrations to 85.6\% at 200, improving on every task. At 100 demonstrations, mean success reaches 77.8\%, compared with 62.2\% for the fixed 50-real-demonstration reference.

\paragraph{Amortized generation cost.}
Using a trained renderer, we estimate the average cost per demonstration over $N$ 30-s, three-view demonstrations of one task:
\begin{equation}
\overline C(N)=\frac{C_{\mathrm{plan}}}{N}+c_{\mathrm{gen}},
\label{eq:cost}
\end{equation}
where $C_{\mathrm{plan}}\approx\$9$ is the one-time GPT-6 Astra programming cost observed in our runs. The recurring image/video cost $c_{\mathrm{gen}}=\$0.765$ assumes 15.3 GPU-minutes per demonstration at \$3/H100 GPU-hour~\cite{runpodpricing}. This assumes 45 GPU-seconds for each of 15 video windows and 30 GPU-seconds for each of three image edits, plus 20\% overhead. Average cost falls from \$9.765 at $N=1$ to \$0.855 at $N=100$ and \$0.774 at $N=1{,}000$. Figure~\ref{fig:scaling}(b) compares these estimates with human teleoperation at \$2.42 per demonstration, based on \$290 per usable data hour including collection and processing overhead~\cite{robotdatacost}.

\subsection{Generalization across Scene Variations}
\label{sec:generalization}
Using a video model as the renderer enables domain randomization through changes to the rendering prompt. We evaluate the resulting policy generalization with $\pi_{0.5}$ on the can-to-tray task under three appearance variations covered during training: a green bottle, a changed background color, and altered lighting. We compare policies trained on simulation domain randomization~\cite{domainrand,robotwin} and DemoFoundry data with language-prompted visual variations. Object placements are randomized on every trial. Over 30 real-robot trials per condition, DemoFoundry achieves 21, 22, and 18 successes, compared with 5, 10, and 2 for Simulation + DR. These results demonstrate stronger visual generalization than standard simulation domain randomization. Appendix C reports the settings and results.

\section{Conclusion}
We present \method, a framework that couples coding agents with generative rendering to automate the synthesis of photorealistic, action-labeled robot demonstrations. Experiments demonstrate improved visual quality, strong action--video consistency, and effective policy learning on real robots.

Remaining challenges include multi-view consistency and drift over repeated autoregressive rollouts, which may be mitigated by joint multi-view generation and memory-based conditioning across video segments. \method offers a path to generating diverse robot interactions on demand with reduced human collection effort, enabling scalable training of more generalizable robot policies.

\label{lastmainpage}
\clearpage
\section*{LLM Usage Statement}
OpenAI Codex assisted with literature search and language polishing. Within DemoFoundry, GPT-6 Astra supports scene reconstruction and task programming, Qwen-Image-2.1 generates reference images, and LTX-2.5 and MiniMax-H3 render demonstration videos, as detailed in Sections 3--4 and the appendix. The authors take full responsibility for the submission.

{\small
\bibliographystyle{ieeenat_fullname}
\bibliography{main}
}
\clearpage
\appendix
\setcounter{table}{0}
\setcounter{figure}{0}
\setcounter{equation}{0}
\renewcommand{\thetable}{S\arabic{table}}
\renewcommand{\thefigure}{S\arabic{figure}}
\renewcommand{\theequation}{S\arabic{equation}}
\renewcommand{\theHtable}{appendix.\arabic{table}}
\renewcommand{\theHfigure}{appendix.\arabic{figure}}
\renewcommand{\theHequation}{appendix.\arabic{equation}}
\twocolumn[{
\begin{center}
{\Large\bfseries DemoFoundry: Toward Scalable Photorealistic Robot Data\\
with Coding Agents and Generative Rendering\par}
\vspace{8pt}
{\large Supplementary Material\par}
\end{center}
\vspace{12pt}
}]
\section{Coding Agent Harness}
\label{sec:supp_harness}
The harness connects GPT-6 Astra to reconstruction, simulation, and evaluation tools. Real--proxy mask comparisons guide scene reconstruction; execution errors and task-completion checks guide task programming. Successful programs are reused across randomized layouts. Table~\ref{tab:supp_harness} summarizes these feedback loops.

\begin{table}[ht]
\centering\small
\caption{\textbf{Execution feedback in the coding agent harness.} Feedback guides scene reconstruction in main Sec.~3.2 and task programming in main Sec.~3.3.}
\label{tab:supp_harness}
\setlength{\tabcolsep}{4pt}
\renewcommand{\arraystretch}{1.12}
\begin{tabularx}{\columnwidth}{@{}YY@{}}
\toprule
\textbf{Source-episode reconstruction} & \textbf{New-task programming} \\
\midrule
\multicolumn{2}{@{}l}{\textit{Execution}} \\
Replay the recorded controls in the reconstructed scene with the original camera calibration. &
Run the agent-written program $p$ in scene $E$ using the robot-control tools in $\mathcal T$. \\
\midrule
\multicolumn{2}{@{}l}{\textit{Feedback and revision}} \\
A silhouette mismatch identifies an object, camera view, and keyframe. Revise the object's geometry, scale, or initial pose and replay the same controls. &
Execution errors or a failed task-completion check guide revisions to $p$. Test the revised program in the simulator. \\
\midrule
\multicolumn{2}{@{}l}{\textit{Acceptance and output}} \\
Keep the real--proxy pair if every visible object's silhouette IoU is at least 0.80 in every checked view and keyframe. Discard it if it still fails after three refinement rounds. &
Reuse a successful program across randomized layouts. For each successful rollout, send $P_i$ to rendering and retain the aligned states $Q_i$ and actions $A_i$. \\
\bottomrule
\end{tabularx}
\end{table}

\paragraph{Tool library $\mathcal T$.}\mbox{}\par

\smallskip
{\noindent\textbf{Reconstruction interfaces.}
Observation and calibration readers establish coordinate and time mappings. Asset-building tools fit task geometry and contact surfaces; robot/gripper checks verify transforms and gripper-opening calibration. Native replay returns simulated trajectories, while image-space comparisons localize alignment errors.\par}

\newpage
\paragraph{Tool library $\mathcal T$ (continued).}\mbox{}\par
\smallskip
{\noindent\textbf{Task-programming interfaces.}
Scene queries return current object poses. Grasp proposals, inverse kinematics, motion planning, and robot-control interfaces support program $p$. Execution returns task-completion feedback and synchronized proxy videos $P$, states $Q$, and actions $A$.\par}

\paragraph{Effect of reconstruction feedback.}
Table~\ref{tab:supp_pair_refinement} evaluates the initial replay and up to three refinement rounds on a fixed, source-balanced set of episodes. An episode passes when every visible object achieves a silhouette IoU $\geq0.80$ across all checked views and keyframes. Pass rates are computed cumulatively over all attempted episodes with fixed calibration, segmentation, and controls, while alignment metrics are evaluated on held-out keyframes. Across three rounds, iterative refinement raises the acceptance rate from 33.1\% to 39.7\%, improves mean mask IoU from 0.517 to 0.601, reduces track error from 1.370\% to 0.956\%, and increases motion correlation to 0.932.

\begin{table}[ht]
\centering\small
\caption{\textbf{Episode acceptance and alignment by refinement round.} Mask IoU measures silhouette overlap. Track error is the mean real/proxy centroid distance as a percentage of the image diagonal. Motion correlation is the Pearson correlation of horizontal and vertical centroid displacements over jointly visible intervals.}
\label{tab:supp_pair_refinement}
\setlength{\tabcolsep}{2.2pt}
\renewcommand{\arraystretch}{1.2}
\begin{tabularx}{\columnwidth}{@{}Ycccc@{}}
\toprule
Reconstruction & Pass rate & Mask & Track error & Motion \\
stage & (\%)$\uparrow$ & IoU$\uparrow$ & (\%)$\downarrow$ & corr.$\uparrow$ \\
\midrule
Initial replay & 33.1 & 0.517 & 1.370 & 0.893 \\
+1 refinement & 35.3 & 0.554 & 1.214 & 0.907 \\
+2 refinements & 36.8 & 0.568 & 1.082 & 0.919 \\
+3 refinements & 39.7 & 0.601 & 0.956 & 0.932 \\
\bottomrule
\end{tabularx}
\end{table}

\section{Renderer and Policy Implementation}
\label{sec:supp_renderer}
\label{sec:supp_policy}
Table~\ref{tab:supp_renderconfig} lists the renderer settings, and Table~\ref{tab:supp_prompt} gives the H3 rendering prompt. Adjacent inference windows overlap by one frame, giving strides of 120 frames for LTX and 157 for H3.

\FullWidthPage{\noindent\begin{minipage}{\textwidth}
\centering\small
\setlength{\tabcolsep}{5pt}
\renewcommand{\arraystretch}{1.08}
\captionof{table}{\textbf{Video renderer training and inference settings.} Paired samples are split into 96\% training, 2\% validation, and 2\% test data. Sizes are width$\times$height.}
\label{tab:supp_renderconfig}
\begin{tabularx}{\textwidth}{@{}p{0.25\textwidth}YY@{}}
\toprule
Setting & LTX renderer & H3 renderer \\
\midrule
Base model & LTX-2.5 22B dev & MiniMax-H3 Ref2VA \\
Adapter initialization & Pretrained Layout-to-Render LoRA v1.0 & Standard LoRA initialization \\
LoRA target modules & Self- and cross-attention (Q/K/V/O), MLP & Attention (Q/K/V/O) \\
LoRA rank / $\alpha$ & 128 / 128 & 32 / 32 \\
Training / inference window & 121 frames & 158 frames \\
AgiBot / RoboMIND size & $512\times384$ & $512\times384$ \\
DROID / RH20T size & $640\times352$ & $640\times352$ \\
Optimizer & AdamW & AdamW \\
Adam $\beta_1,\beta_2$ / $\epsilon$ & $(0.9,0.999)$ / $10^{-8}$ & $(0.9,0.95)$ / $10^{-8}$ \\
Learning-rate schedule & Linear $2\times10^{-4}\!\to\!2\times10^{-5}$ over 3,000 updates, then constant & 100-update warmup to $5\times10^{-5}$, then constant \\
Weight decay / global norm clip & 0.01 / 1.0 & 0.01 / 1.0 \\
Batch size (clips per update) & 32 & 32 \\
Adapter / optimizer precision & BF16 / FP32 master weights and moments & FP32 / FP32 \\
Checkpoint update (main Table 2) & 37,000 & 7,000 \\
\midrule
Denoising steps & 30 & 50 \\
Sampler & Native LTX2Scheduler with Euler & Native H3 sampler \\
Guidance & CFG scale 3 & Native guidance-distilled sampling \\
Default inference seed & 42 & 42 \\
\bottomrule
\end{tabularx}
\end{minipage}

\medskip
\noindent\begin{minipage}{\textwidth}
\centering\small
\captionof{table}{\textbf{Six-part H3 rendering prompt.} The native configuration identifies the RGB reference as \texttt{<Image 1>} and the proxy as \texttt{<Video 1>}. The task instruction $\ell$ is used for programming and policy learning. Appearance edits modify the descriptive and retention clauses.}
\label{tab:supp_prompt}
\setlength{\tabcolsep}{5pt}
\renewcommand{\arraystretch}{1.13}
\begin{tabularx}{\textwidth}{@{}p{0.24\textwidth}Y@{}}
\toprule
Section & Prompt text \\
\midrule
\texttt{subject\_definitions} & \textless{}Video 1\textgreater{} is a simplified simulation rendering of the robot and task geometry, from the same camera as the target video.\newline \textless{}Image 1\textgreater{} is the appearance reference image for the first frame of [Shot 1]. \\
\texttt{summary} & [keyframe completion + reference generation] The target video begins from \textless{}Image 1\textgreater{} and follows the frame-by-frame structure and motion of \textless{}Video 1\textgreater{}. A photographic recording of a robot manipulation scene, matching the supplied first frame and following the supplied structure sequence. \\
\texttt{retention\_analysis} & \textless{}Video 1\textgreater{} (frame-by-frame layout, shapes, character poses and motion, camera): attribute\_transfer - the layout, shapes, poses, motion and camera are followed exactly; materials, colours and lighting are not given by \textless{}Video 1\textgreater{} and are the target video's own; the simplified simulation materials and plain background of \textless{}Video 1\textgreater{} is not shown.\newline \textless{}Image 1\textgreater{} ([Shot 1] first frame): fully\_preserved - the composition, subjects, materials, colours and lighting of \textless{}Image 1\textgreater{} are the target video's first frame. \\
\texttt{detailed\_description} & [Shot 1] The shot begins from \textless{}Image 1\textgreater{}. The target is a continuous photographic recording of the robot workspace shown in the first frame. Preserve the robot's visible identity, link shapes, gripper, table surface, task objects, room background, materials, colours and lighting. Follow only the poses, motion, occlusion and object trajectories visible in the structure sequence in playback order. Keep the camera viewpoint and any camera motion consistent with that sequence. Maintain the actual photographic background throughout the clip. Preserve object count and connected robot geometry. Do not introduce a new action or an unseen object. Keep textures and illumination stable as the robot moves. \\
\texttt{overall\_soundscape} & N/A \\
\texttt{non\_diegetic\_music} & N/A \\
\bottomrule
\end{tabularx}
\end{minipage}

}

\twocolumn[{
\noindent\begin{minipage}{\textwidth}
\centering\small
\renewcommand{\arraystretch}{1.12}
\setlength{\tabcolsep}{5pt}
\captionof{table}{\textbf{Policy training and deployment settings.} Settings are fixed across data conditions.}
\label{tab:supp_policyconfig}
\begin{tabularx}{\textwidth}{@{}p{0.28\textwidth}YY@{}}
\toprule
Setting & $\pi_{0.5}$ & Diffusion Policy \\
\midrule
Initialization & \texttt{pi05\_base} & --- \\
Trainable parameters & Language/action LoRA & All policy parameters \\
Optimizer & AdamW & AdamW \\
Adam $(\beta_1,\beta_2)$ & $(0.9,0.95)$ & $(0.95,0.999)$ \\
Learning rate & $5\times10^{-5}\!\to\!5\times10^{-6}$, cosine decay, 500-step warmup & $10^{-4}$ \\
Weight decay / gradient clip & $10^{-10}$ / 1.0 & $10^{-6}$ / 1.0 \\
Updates / batch size & 20,000 / 32 & 50,000 / 64 \\
Action horizon & 50 steps & 8 steps \\
Control frequency & 25\,Hz & 25\,Hz \\
\bottomrule
\end{tabularx}
\end{minipage}

\paragraph{Policy interface.}
The $\pi_{0.5}$ interface maps three camera views and 16 joint/gripper coordinates to its 32-dimensional representation. States and actions use training-set quantile normalization. Arm targets are joint offsets, and gripper targets are absolute values.

\medskip
\section{Domain Randomization and Generalization}
\label{sec:supp_generalization}
\noindent\begin{minipage}{\textwidth}
\centering
\includegraphics[width=\textwidth]{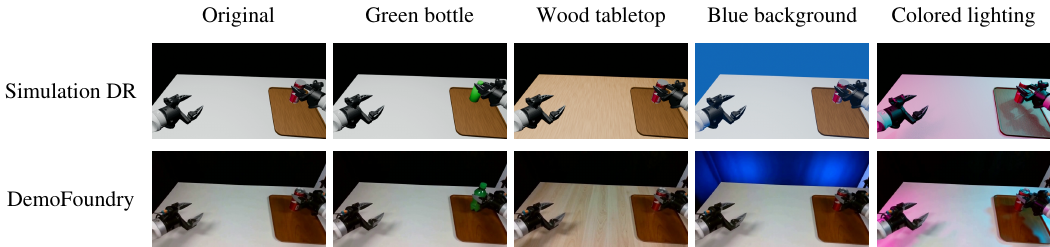}
\captionof{figure}{\textbf{Simulation randomization and prompt-controlled rendering.} The leftmost column shows the original appearance. Subsequent columns compare a green bottle, a wood-grain tabletop, a blue background, and cyan/magenta lighting. The top row shows domain randomization in simulation; the bottom row shows appearance variations generated by changing the rendering prompt in DemoFoundry.}
\label{fig:supp_dr}
\end{minipage}

\vspace{14pt}
}]
\paragraph{Training data.}
Figure~\ref{fig:supp_dr} illustrates two ways to vary the appearance of a demonstration. For Simulation + DR, we manually configure object and tabletop materials, background colors, and lighting in the simulator. With DemoFoundry, editing the appearance prompt lets the video renderer produce photorealistic variations from the same RGB reference and proxy video.

\paragraph{Real-robot evaluation.}
We evaluate $\pi_{0.5}$ on the can-to-tray task under three target visual variations covered by the respective randomization pipelines: a green bottle, a changed background color, and altered lighting. Object placements are randomized on every trial, with 30 real-robot trials per policy per condition.

\newpage
\begin{table}[h!]
\centering\footnotesize
\renewcommand{\arraystretch}{1.12}
\caption{\textbf{Generalization on the can-to-tray task.} Entries report $\pi_{0.5}$ successes out of 30 real-robot trials per condition.}
\label{tab:supp_generalization}
\setlength{\tabcolsep}{3pt}
\begin{tabularx}{\columnwidth}{@{}Y*{3}{>{\centering\arraybackslash}X}@{}}
\toprule
Training data & Green bottle & Background color & Lighting \\
\midrule
Simulation + DR & 5/30 & 10/30 & 2/30 \\
\textbf{DemoFoundry} & \textbf{21/30} & \textbf{22/30} & \textbf{18/30} \\
\bottomrule
\end{tabularx}
\end{table}
\medskip

As reported in Table~\ref{tab:supp_generalization}, DemoFoundry outperforms Simulation + DR across all three settings, achieving 18/30 versus 2/30 successes under altered lighting.

\FullWidthPage{
\section{Real--Proxy Video Pairs}
\label{sec:supp_real_proxy_pairs}
\noindent\begin{minipage}{\textwidth}
\centering
\includegraphics[width=\textwidth,height=0.85\textheight,keepaspectratio]{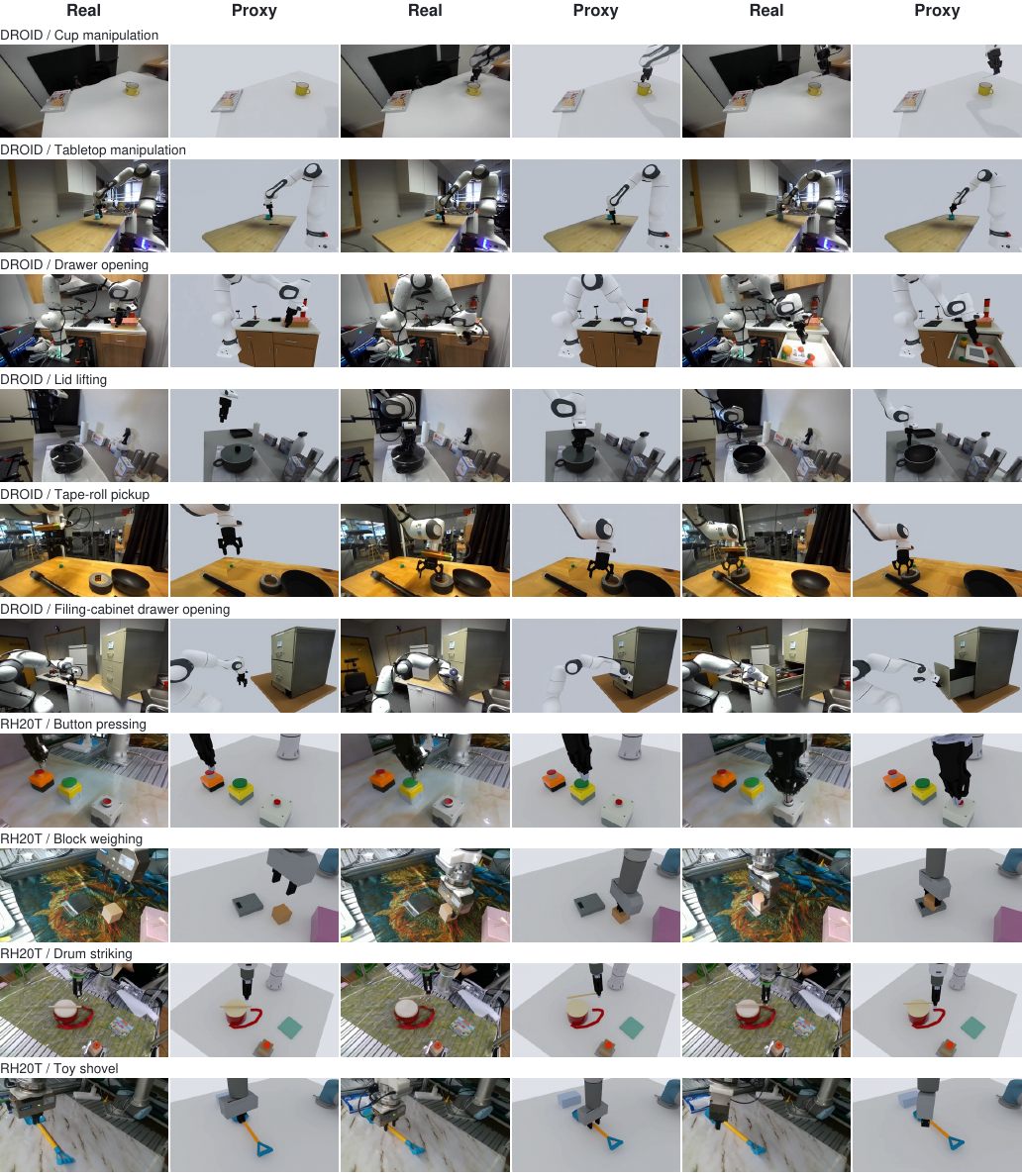}
\captionof{figure}{\textbf{Real--proxy pairs from DROID and RH20T.} The first six rows show DROID episodes and the last four show RH20T episodes. Each row presents three time-aligned real--proxy pairs in chronological order.}
\label{fig:supp_real_proxy_pairs}
\end{minipage}
}

\FullWidthPage{
\noindent\begin{minipage}{\textwidth}
\centering
\includegraphics[width=\textwidth,height=0.90\textheight,keepaspectratio]{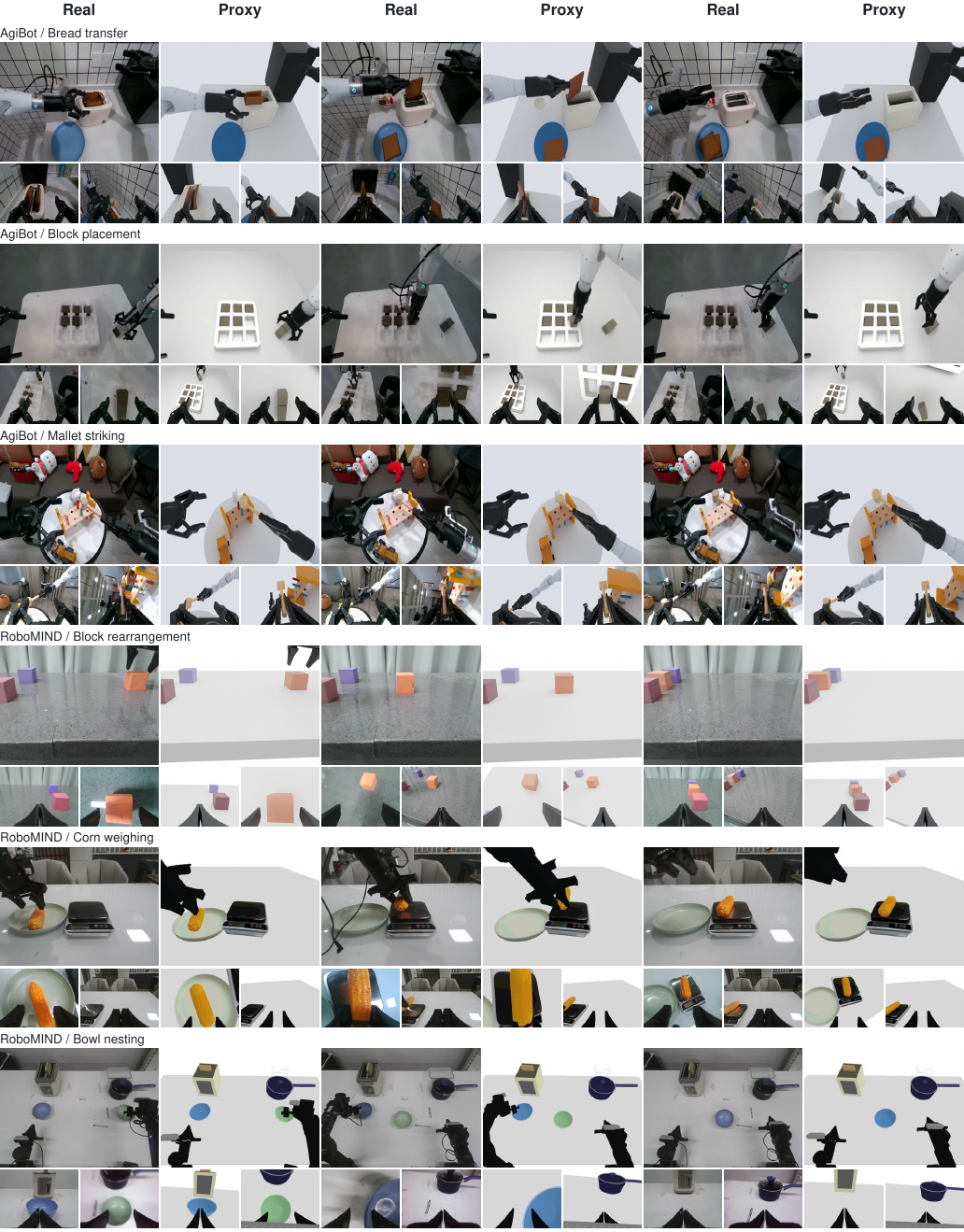}
\captionof{figure}{\textbf{Real--proxy pairs from AgiBot and RoboMIND.} The first three rows show AgiBot episodes and the last three show RoboMIND episodes. Each row presents three time-aligned real--proxy pairs in chronological order, with head/front and wrist views grouped in each cell.}
\label{fig:supp_multiview_pairs}
\end{minipage}
}

\FullWidthPage{
\section{Temporal Generation Comparisons}
\label{sec:supp_temporal}
Figures~\ref{fig:supp_focused_1}--\ref{fig:supp_focused_4} present qualitative temporal comparisons on held-out episodes across the four source datasets, where H3 and LTX denote the released MiniMax-H3 Ref2VA and LTX-2.5 Layout-to-Render models, respectively.

\medskip
\noindent\begin{minipage}{\textwidth}
\raggedright
\noindent\hspace*{-3pt}\includegraphics[width=0.98\textwidth]{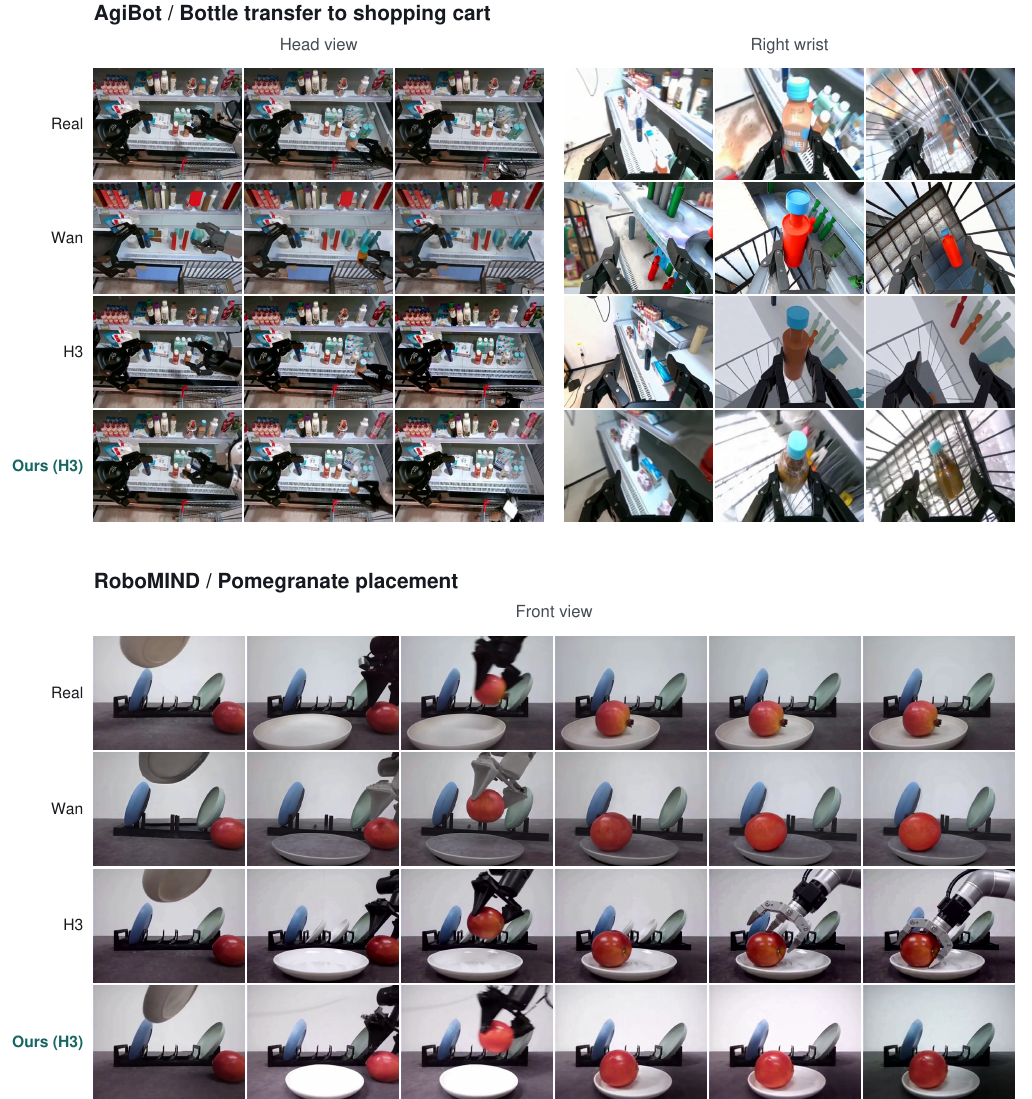}
\captionof{figure}{\textbf{MiniMax-H3 rendering of bottle transfer and fruit placement.} Released H3 retains proxy-like colors and introduces a gripper absent from the real fruit-placement frames. Paired adaptation improves scene appearance and gripper--object correspondence.}
\label{fig:supp_focused_1}
\end{minipage}
}

\FullWidthPage{
\noindent\begin{minipage}{\textwidth}
\raggedright
\noindent\hspace*{-3pt}\includegraphics[width=0.98\textwidth]{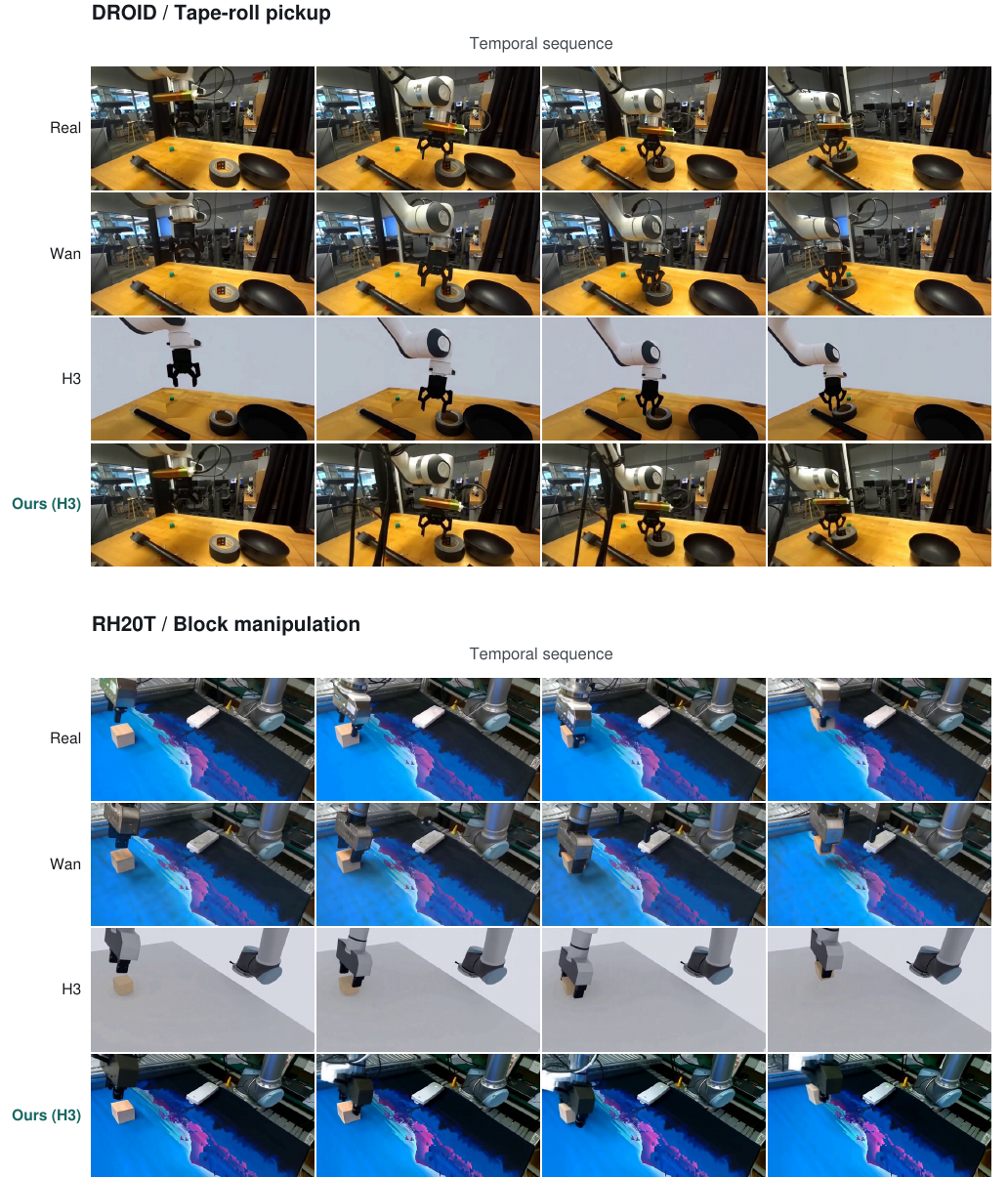}
\captionof{figure}{\textbf{MiniMax-H3 rendering of tape-roll pickup and block manipulation.} Released H3 retains simplified simulation backgrounds, whereas paired adaptation recovers the wooden tabletop and blue work mat.}
\label{fig:supp_focused_2}
\end{minipage}
}

\FullWidthPage{
\noindent\begin{minipage}{\textwidth}
\raggedright
\noindent\hspace*{-3pt}\includegraphics[width=0.98\textwidth]{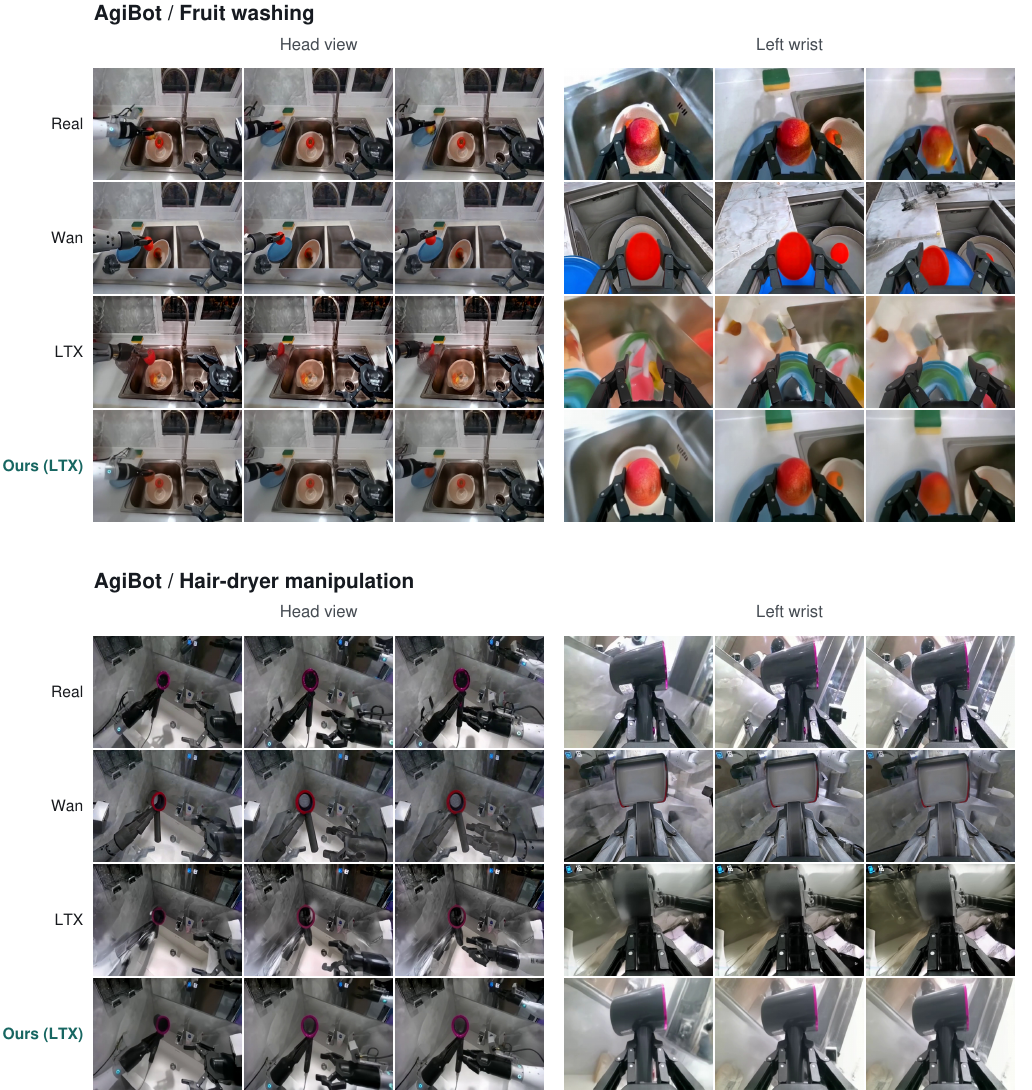}
\captionof{figure}{\textbf{LTX rendering of fruit washing and hair-dryer manipulation.} Released LTX distorts object shapes in the wrist views, whereas paired adaptation better preserves fruit appearance and hair-dryer shape.}
\label{fig:supp_focused_3}
\end{minipage}
}

\FullWidthPage{
\noindent\begin{minipage}{\textwidth}
\raggedright
\noindent\hspace*{-3pt}\includegraphics[width=0.98\textwidth]{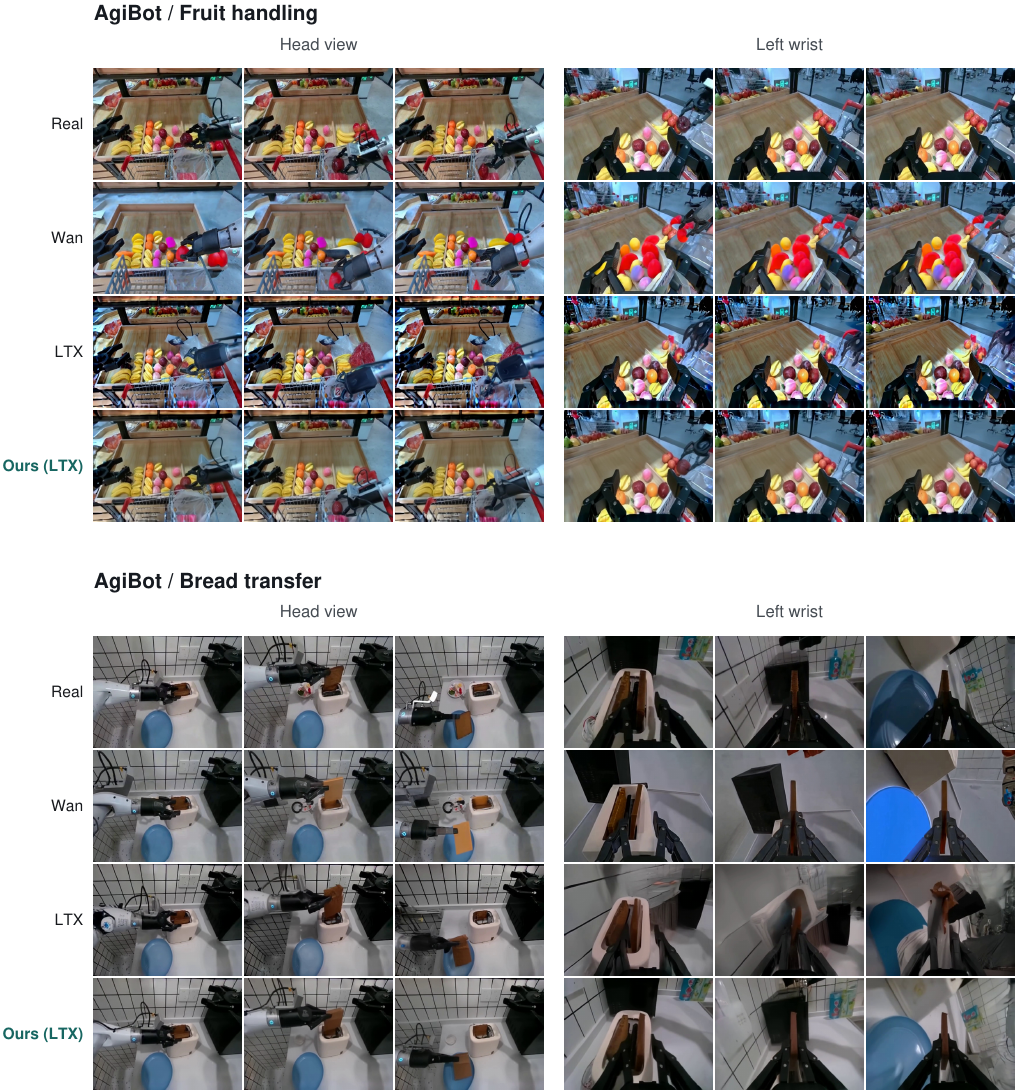}
\captionof{figure}{\textbf{LTX rendering of fruit handling and bread transfer.} Released LTX introduces texture artifacts and blurs the bread and gripper in wrist views, whereas paired adaptation produces clearer object boundaries and more consistent scene appearance.}
\label{fig:supp_focused_4}
\end{minipage}
}

\end{document}